\documentclass{article}
\usepackage{spconf,amsmath,graphicx}

\usepackage[linesnumbered,ruled,vlined]{algorithm2e}
\usepackage{algpseudocode}
\usepackage{amsmath}

\usepackage{booktabs}       

\usepackage{url}

\title{CIF: Continuous Integrate-and-Fire for End-to-End Speech Recognition}
\name{Linhao Dong$^{\star \dagger}$ \qquad Bo Xu$^{\star}$
\thanks{This work is supported by the National Key Research and Development Program of China under No.2018YFB1005104 and Beijing Municipal Science and Technology Project under No.Z181100008918017.}}

\address{
$^{\star}$Institute of Automation, Chinese Academy of Sciences, China\\
$^{\dagger}$University of Chinese Academy of Sciences, China\\
\small \tt \{donglinhao2015, xubo\}@ia.ac.cn}

\begin{document}
\ninept
\renewcommand{\baselinestretch}{0.8812} \normalsize

%
\maketitle
\begin{abstract}
  In this paper, we propose a novel soft and monotonic alignment mechanism used for sequence transduction. It is inspired by the integrate-and-fire model in spiking neural networks and employed in the encoder-decoder framework consists of continuous functions, thus being named as: Continuous Integrate-and-Fire (CIF). Applied to the ASR task, CIF not only shows a concise calculation, but also supports online recognition and acoustic boundary positioning, thus suitable for various ASR scenarios. Several support strategies are also proposed to alleviate the unique problems of CIF-based model. With the joint action of these methods, the CIF-based model shows competitive performance. Notably, it achieves a word error rate (WER) of 2.86\% on the test-clean of Librispeech and creates new state-of-the-art result on Mandarin telephone ASR benchmark.
\end{abstract}

\section{INTRODUCTION}
\label{sec:intro}
Automatic speech recognition (ASR) system is undergoing an exciting pathway to be more simplified and accurate with the spring up of various end-to-end models. Among them, the attention-based model \cite{chorowski2015attention, chan2016listen}, which builds a soft alignment between each decoder step and every encoder step, not only shows a performance advantage in comparison with other end-to-end models \cite{prabhavalkar2017comparison}, but also successfully challenges the dominance of HMM-LSTM Hybrid system in ASR \cite{chiu2018state}. However, despite the superiority of accuracy, such attention-based model often encounters incompetent scenarios in real ASR application: 1) it cannot support online (or streaming) recognition since it need refer to the entire encoded sequence; 2) it cannot well time-stamp the recognition result since it's not frame-synchronous. Besides, attending to every encoder steps is bound to bring a mass of unnecessary computations on steps that are acoustically irrelevant to the decoding step. Focusing on solving above problems, we aim at seeking a soft alignment which not only performs an efficient monotonic calculation but also locates acoustic boundaries. And we find inspirations from the integrate-and-fire model \cite{lapicque1907recherches, abbott1999lapicque}.

Integrate-and-fire is one of the earliest models in spiking neural networks (SNNs), which are more bio-plausible and known as the next generation of neural networks \cite{Maass1997Networks}. The integrate-and-fire neuron operates using spikes, which are discrete events that take place at points in time. Specifically, it forwardly integrates the stimulations in the input signal (e.g. spike train), and its membrane potential changes accordingly. When the potential reaches a specific threshold, it fires a spike that will stimulate other neurons, and its potential is reset. It's not hard to find that: 1) such integrate-and-fire process is strictly monotonic; 2) the fired spikes could be used to represent the events that locate an acoustic boundary. By transferring the idea of integrate-and-fire to the end-to-end ASR, we could imagine such an alignment mechanism: it forwardly integrates the information in acoustic signals, once a boundary is located, it instantly fires the integrated acoustic information for further recognition. And the difficulty of achieving it lies in how to simulate the process of integrate-and-fire using continuous functions that support back-propagation.

\begin{figure}
  \centering
  \includegraphics[width=\linewidth]{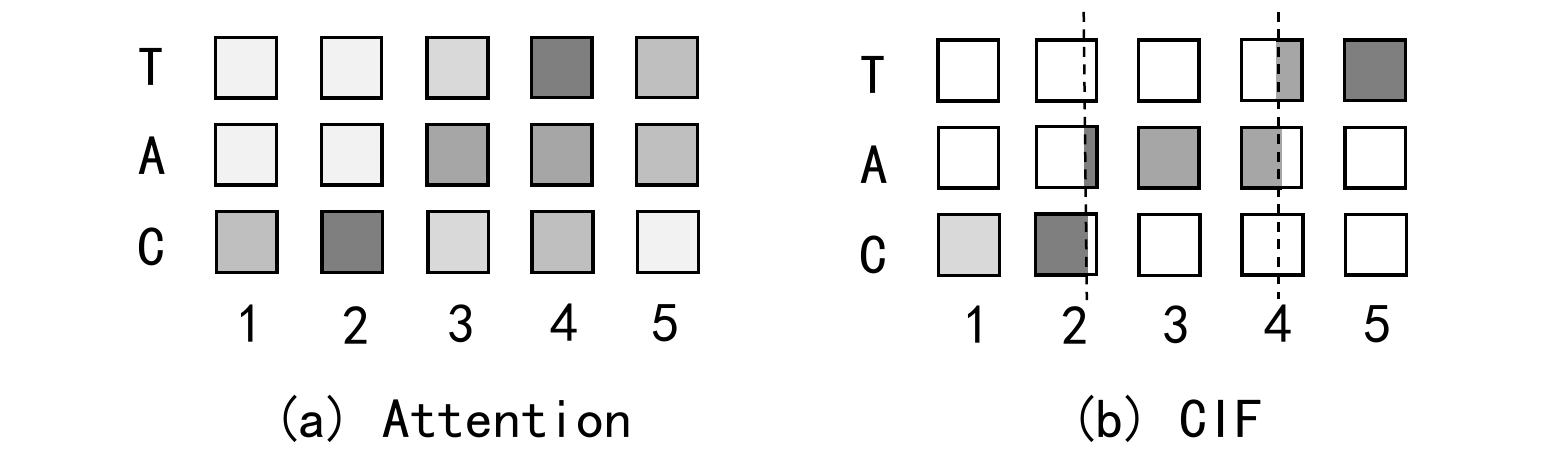}
  \vspace{-13.5pt}
  \caption{Illustration of the attention alignment and our proposed CIF alignment on an encoded utterance of length 5 and labelled as "CAT". The shade of gray in each square represents the weight of each encoder step involved in the calculation of decoding labels. The vertically dashed line in (b) represents the located acoustic boundary, and the weight of the boundary step is divided into two parts used for the calculation of the two adjacent labels, respectively.}
  \label{fig:alignment}
  \vspace{-13.5pt}
\end{figure}

In this paper, we propose Continuous Integrate-and-Fire (CIF), a novel soft and monotonic alignment employed in the encoder-decoder framework. At each encoder step, it receives the vector representation of current encoder step and a corresponding weight that scales the amount of information contained in the vector. Then, it forwardly accumulates the weights and integrates the vector information until the accumulated weight reaches a threshold, which means an acoustic boundary is located. At this point, the acoustic information of current encoder step is shared by two adjacent labels, thus CIF divides the information into two part: the one for completing the integration of current label and the other for the next integration, which mimics the processing of the integrate-and-fire model when it fires at some point during the period of a encoder step. Then, it fires the integrated acoustic information to the decoder to predict current label. Such process is sketched in Fig.\ref{fig:alignment} (b) and is performed till to the end of the encoded sequence.

We also present several supporting strategies to refine the performance of CIF-based model, including: 1) a scaling strategy to solve the problem of unequal length between the predicted labels and the targeted labels in the cross-entropy training; 2) a quantity loss to supervise the model to predict the quantity of labels closer to the targets; 3) a tail handling method to process the residual information at the end of inference. With the joint action of these methods, our CIF-based model shows impressive performance on multiple ASR datasets covering different languages and speech types.

\section{RELATION TO PRIOR WORK}
\label{sec:prior}
Several prior works have also studied the soft and monotonic alignment in end-to-end ASR models. \cite{hou2017gaussian, tjandra2017local, Merboldt2019} assumes the alignment to be a forward-moving window that fits gaussian distribution \cite{hou2017gaussian, tjandra2017local} or even heuristic rule \cite{Merboldt2019}, where the center and width of the window are predicted by its decoder state. Comparing with them, CIF neither follows a given assumption nor uses the state of the decoder, thus encouraging more pattern learning from the acoustic data.

Besides, CIF provides a concise calculation process by conducting the locating and integrating at the same time, rather than \cite{chiu2017monotonic, fan2018online} which need two separate steps of first using a hard monotonic attention to decide when to stop and then performing soft attention to calculate, also rather than \cite{moritz2019triggered} which needs a CTC trained model to conduct pre-partition before the attention decoding.

In \cite{li2019end}, Li al. present the important Adaptive Computation Steps (ACS) algorithm whose motivation is to dynamically decide a block of frames to predict a linguistic output. In comparison, CIF holds a different motivated perspective --- `integrate-and-fire', and models at a finer time granularity to process the firing phenomenon widely existed inside the encoded frames. Besides, the processing of CIF ensures the full utilization of acoustic information and lays a foundation for the effective application of its supporting strategies, which are what the ACS (that has a huge performance gap from a HMM-DNN model) lacks of.

\section{Method}
\label{sec:method}

\begin{figure}
  \centering
  \includegraphics[width=\linewidth]{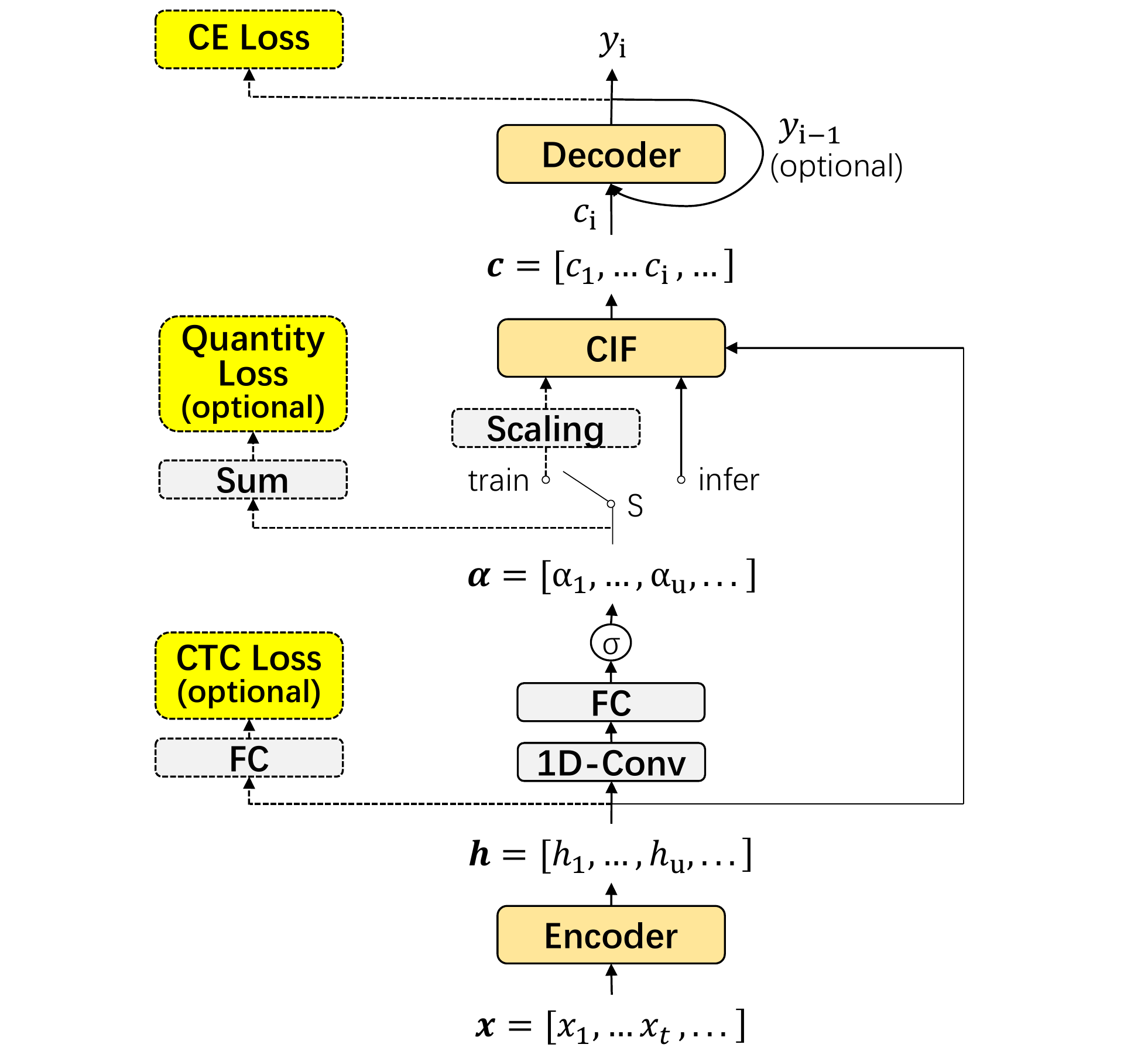}
  \vspace{-13.5pt}
  \caption{The architecture of our CIF-based model used for the ASR task. Operations in the dashed rectangles are only applied in the training stage. The switch (S) before the CIF module connects the left in the training stage and the right in the inference stage.}
  \label{fig:framework}
  \vspace{-13.5pt}
\end{figure}

\subsection{Continuous Integrate-and-Fire}
\label{subsec:cif}
\begin{figure}
  \centering
  \includegraphics[width=225pt]{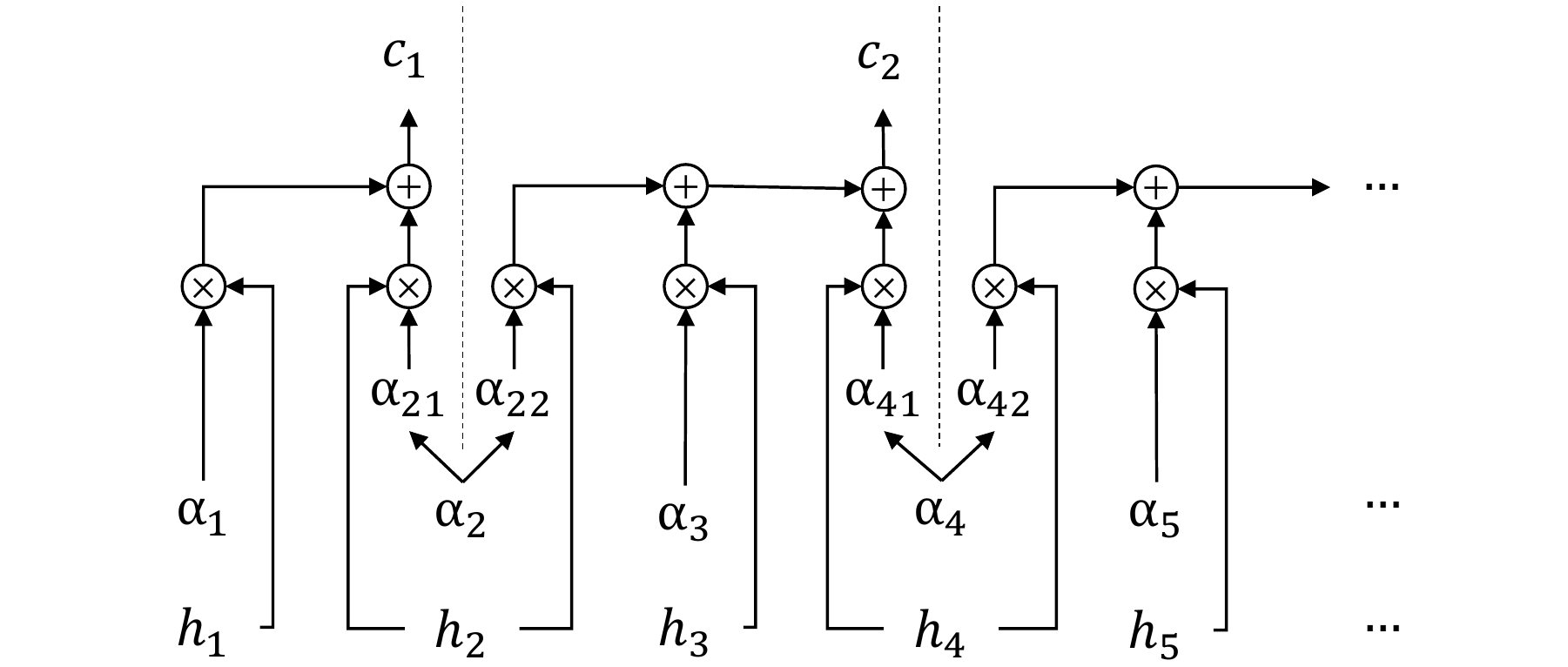}
  \vspace{-4.5pt}
  \caption{Illustration of the calculation of CIF on an encoded sequence $\mathbf{h} = (h_1, h_2, h_3, h_4, h_5, \ldots)$ with predicted weights $\boldsymbol{\alpha} = (0.2, 0.9, 0.6, 0.6, 0.1, \ldots)$. The integrated embedding $c_1=0.2*h_1+0.8*h_2$, $c_2=0.1*h_2+0.6*h_3+0.3*h_4$. }
  \label{fig:cif}
  \vspace{-13.5pt}
\end{figure}

Continuous Integrate-and-Fire (CIF) is a soft and monotonic alignment employed in the encoder-decoder framework. As shown in Fig.\ref{fig:framework}, CIF connects the encoder and decoder. At each encoder step $u$, it receives two inputs: 1) current output (state) of encoder: $h_u$; 2) current weight: $\alpha_u$, which scales the amount of information contained in $h_u$. Then, it forwardly accumulates the received weights and integrates the received states (using the form of `weighted sum') until the accumulated weight reaches a given threshold $\beta$, which means an acoustic boundary is located. At this point, the information of current encoder step is shared by current label $y_i$ and the next label, thus CIF divides current weight $\alpha_u$ into two part: the one is used to fulfill the integration of current label $y_i$ by building a complete distribution (whose sum of weights is 1.0) on relevant encoder steps, the other is used for the integration of next label. After that, it fires the integrated embedding $c_i$ (as well as the context vector) to the decoder to predict the corresponding label $y_i$. The above process is roughly presented in Fig.\ref{fig:cif} and is performed till to the end of encoded sequence. The complete algorithm is detailed in Algorithm 1, where the threshold $\beta$ is recommended to be 1.0.

\begin{algorithm}
\caption{Continuous Integrate-and-Fire (CIF)}
\LinesNumbered
\KwIn{The outputs of encoder $\mathbf{h} = (h_1, \ldots, h_u, \ldots, h_U)$ and the corresponding weights $\boldsymbol{\alpha} = (\alpha_1, \ldots, \alpha_u, \ldots, \alpha_U)$, the threshold $\beta$;}
\KwOut{The integrated embeddings (corresponding to the output labels): $\mathbf{c} = (c_1, \ldots, c_i, \ldots, c_S)$;}
\textbf{Initialize} $i = 1$, initial accumulated weight $\alpha_0^a = 0$, initial accumulated state $h_0^a = \mathbf{0}$; \\
\For{  $u=1$; $u<=U$; $u++$  }
{
    // calculate current accumulated weight; \\
    $\alpha_u^a = \alpha_{u-1}^a + \alpha_u$; \\
    \If{ $\alpha_u^a < \beta $ }
    {
        // no boundary is located; \\
        $h_u^a = h_{u-1}^a + \alpha_u * h_u $;
    }
    \Else
    {
        // a boundary is located; \\
        // $\alpha_u$ is divided into two part, the first part $\alpha_{u1}$ is used to fulfill the integration of current label $y_i$; \\
        $\alpha_{u1} = 1 - \alpha_{u-1}^a$; \\
        $c_i = h_{u-1}^a + \alpha_{u1} * h_u$; \\
        i++; \\
        // The other part $\alpha_{u2}$ is used for the next integration; \\
        $\alpha_u^a = \alpha_{u2} = \alpha_u - \alpha_{u1}$; \\
        $h_u^a = \alpha_{u2} * h_u$; \\
    }
}
\textbf{return} $\mathbf{c} = (c_1, \ldots, c_i, \ldots, c_S)$;
\end{algorithm}
\vspace{-13.5pt}

\subsection{Supporting Strategies for CIF-based Model}
\label{subsec:supporting}
We also present some support strategies for the CIF-based model to alleviate its unique problems during training and inference:

\textbf{Scaling Strategy}: In the training, the length $S$ of the produced integrated embeddings $\mathbf{c}$ may differ from the length $\tilde{S}$ of targets $\tilde{\mathbf{y}}$, thus bringing difficulties to the cross-entropy training that better to be `one-to-one'. To solve it, we propose a scaling strategy, which multiplies the calculated weights $\boldsymbol{\alpha}=(\alpha_1, \alpha_2, ..., \alpha_U)$ by a scalar $\frac{\tilde{S}}{\sum_{u=1}^U{\alpha_u}}$ to generate the scaled weights $\boldsymbol{\alpha{'}}=(\alpha_1{'}, \alpha_2{'}, ..., \alpha_U{'})$ whose sum is equal to $\tilde{S}$, thus teacher-forcing CIF to produce $\mathbf{c}$ with length $\tilde{S}$ and driving more effective training.

\textbf{Quantity Loss}: We also present an optional loss function to supervise the CIF-based model to predict the quantity of integrated embeddings closer to the quantity of targeted labels. We term it as quantity loss $\mathcal{L}_{QUA}$, which is defined as $\left| \sum_{u=1}^U\alpha_{u} - \tilde{S} \right|$, where $\tilde{S}$ is the length of the targets $\tilde{\mathbf{y}}$. By providing the quantity constraints, this loss not only promotes the learning of acoustic boundary positioning, but also alleviates the performance degradation after removing the scaling strategy in the inference.

\textbf{Tail Handling}: In the inference, the tail leaves some useful information that is not enough to trigger one firing. Directly discarding this information causes the incomplete results (e.g. incomplete words in ASR) at the tail. To alleviate such problem, we utilize a rounding method which makes an additional firing if the residual weight is greater than 0.5 during inference. Besides, we also introduce a label \textless{EOS}\textgreater\ to the tail of target sequence to teach the model to predict the end of sentence and more importantly, provide cache.

\subsection{Model Structure}
\label{sec:model}
Fig.\ref{fig:framework} shows the architecture of our CIF-based model used for ASR but lacks some details of the model structure. Here, we give these details and introduce some new characteristics of the CIF-based model:

\textbf{Encoder}: Our encoder follows the encoder structure in \cite{dong2019self}, which uses a two-layer convolutional front-end followed by a pyramid self-attention networks (SANs) and reduces the time resolution to 1$/$8. Forward encoding for online recognition is achieved by applying the chunk-hopping mechanism in \cite{dong2019self}. As an aside, adjusting the encoding resolution enables CIF suitable for various tasks, e.g. we could use up-sampling to generate longer encoded sequence than outputs to make CIF apply to text-to-speech (TTS), etc.

To calculate the weight $\alpha_u$ corresponding to each encoded output $h_u$, we pass a window centered at $h_u$ (e.g. $[h_{u-1},h_u,h_{u+1}]$) to a 1-dimensional convolutional layer and then a fully connected layer with one output unit and a sigmoid activation, where the convolutions can be replaced by other neural networks.

\textbf{Decoder}: Two versions of decoder are introduced in this work: the one is an autoregressive decoder, which follows the decoder structure in \cite{dong2019self}. Specifically, it first projects the concatenation of the embedding ($e_{i-1}$) of the previous label and the previously integrated embedding ($c_{i-1}$) as the input of SANs. Then, it concatenates the output of SANs ($o_i$) and currently integrated embedding ($c_i$) and then projects the concatenation to obtain the logit.

The other is a non-autoregressive decoder, which directly passes the currently integrated embedding ($c_i$) to the SANs to get the output ($o_i$) that is then projected to get the logit. Compared with the autoregressive decoder, it has higher computational parallelization and could provide inference speedups for the offline ASR where the integrated embeddings can be calculated by CIF at once.


\textbf{Loss Functions}: In the training, the SAN-based encoder and decoder provide high parallelization to the teacher-forcing learning of CIF-based model, where the encoding, the CIF calculation (which is lightweight since it just performs the weighted calculation and has no trainable parameters) and the decoding are performed in order. To further boost the model learning, in addition to the cross-entropy loss $\mathcal{L}_{CE}$, two optional auxiliary loss functions are  applied: one of them is the quantity loss $\mathcal{L}_{QUA}$ in section \ref{subsec:supporting}, the other is the CTC loss $\mathcal{L}_{CTC}$, which is applied on the encoder (similar to \cite{kim2017joint}) and addresses the left-to-right acoustic encoding.
When using these two optional loss, our model is trained under the loss $\mathcal{L}$ as follows:
\begin{equation}
  \mathcal{L} = \mathcal{L}_{CE} + \lambda_1 \mathcal{L}_{CTC} + \lambda_2 \mathcal{L}_{QUA}
  \label{eqe}
\end{equation}
where $\lambda_1$ and $\lambda_2$ are tunable hyper-parameters. The importance of the two optional loss functions are explored in section \ref{subsec:ablat}.

\textbf{LM Incorporation}: In the inference, we first perform beam search on the output distributions predicted by the decoder, then use a SAN-based language model (LM) to perform second-pass rescoring as \cite{chiu2018state}, which determines the final transcript $y^*$ as follows:
\begin{equation}
  \mathbf{y^*} = \mathop{\arg\max}_{\mathbf{y} \in \text{NBest}(\mathbf{x},N)} (\text{ log } P(\mathbf{y}|\mathbf{x}) + \gamma \text{ log }P_{LM}(\mathbf{y}))
  \label{eq8}
\end{equation}
where $\gamma$ is a tunable hyper-parameter, $\text{NBest}(\mathbf{x},N)$ is the hypotheses produced by the CIF-based model via beam search with size $N$.

\section{EXPERIMENTAL SETUP}
\label{sec:exp}
We experiment on three public ASR datasets including the popular English read-speech corpus (Librispeech \cite{panayotov2015librispeech}), current largest Mandarin read-speech corpus (AISHELL-2 \cite{du2018aishell}) and the Mandarin telephone ASR benchmark (HKUST \cite{liu2006hkust}). For Librispeech, we use all the train data (960 hours) for training, mix the two development sets for validation, use the two test sets for evaluation, and use the separately prepared language model (LM) data for the training of LM. For AISHELL-2, we use all the train data (1000 hours) for training, mix the three development sets for validation and use the three test sets for evaluation. For HKUST, we use the same training ($\sim$168 hours), validation and evaluation set as \cite{dong2019self}. The training of LM on AISHELL-2 and HKUST uses the text from respective training set.

We extract input features using the same setup as \cite{dong2019self} for all datasets. Speed perturbation \cite{ko2015audio} with fixed $\pm$ 10\% is applied for all training datasets. The frequency masking and time masking in \cite{park2019specaugment} with $F=8$, $m_F=2$, $T=70$, $m_T=2$, $p=0.2$ are applied to all models except the base model on Librisppech. We use the BPE \cite{sennrich2016neural}
toolkit generating 3722 word pieces for Librispeech by merging 7500 times on its training text, plus three special labels: the blank \textless{BLK}\textgreater, the end of sentence \textless{EOS}\textgreater\ and the pad \textless{PAD}\textgreater, the number of output labels is 3725 for Librispeech. We collect the characters and markers from the training text of AISHELL-2 and HKUST, respectively. Plus the three special labels, we generate 5230 output labels for AISHELL-2 and 3674 output labels for HKUST .

We implement our model on TensorFlow \cite{abadi2016tensorflow}. For the self-attention networks (SANs) in our model, we use the structure in \cite{dong2019self} and set $h=4$, $d_{model}=640$, ${d_{ff}}=2560$ for the two Mandarin datasets, and change ($d_{model}$, $d_{ff}$) to (512, 2048), (1024, 4096) for the base, big model on Librispeech, respectively. For the encoder, we use the same configures as \cite{dong2019self}, where $n$ in the pyramid structure is all set to 5. The chunk-hopping \cite{dong2019self} for forward encoding uses the chunk size of 256 (frames) and the hop size of 128 (frames). For the 1-dimensional convolutional layer that predicts weights, the number of filters is set to $d_{model}$, and the window width is all set to 3 except the base model on Librispeech is set to 5. Besides, layer normalization \cite{ba2016layer} and a ReLU activation are applied after this convolution. For CIF, we set the threshold $\beta$ to 0.9 for all models to allow the possible firing after a single step. But it may produce few negative weight after dividing weight, which is non-intuitive in weighted sum calculation. Here, we recommend $\beta=1.0$, which calculates intuitively and performs slightly better than $\beta=0.9$ in our later experiments. For the decoder, the number of SANs is all set to 2 except the base model on Librispeech is set to 3. The loss hyper-parameter $\lambda_1$ is set to 0.5 for two Mandarin datasets and to 0.25 for Librispeech, $\lambda_2$ is all set to 1.0. The LM uses SANs with $h=4$, $d_{model}=512$, $d_{ff}=2048$, and the number of SAN layers is set to 3, 6, 20 for HKUST, AISHELL-2 and Librispeech, respectively.

In the training,
we only apply dropout to the SANs, whose attention dropout and residual dropout are all set to 0.2 except the base model on Librispeech that is set to 0.1. We use the uniform label smoothing in \cite{chorowski2016towards} and set it to 0.2 for both of the CIF-based model and the LM. Scheduled Sampling \cite{bengio2015scheduled} with a constant sampling rate of 0.5 is applied on two Mandarin datasets.
In the inference, we use beam search with size 10. The hyper-parameter $\gamma$ for LM rescoring is set to 0.1, 0.2, 0.9 for HKUST, AISHELL-2 and Librispeech, respectively. All experimental results are averaged over 3 runs.

We display the aligned results (the located boundaries) of CIF on \url{https://linhodong.github.io/cif_alignment/}.

\section{RESULTS}
\label{sec:res}

\subsection{Results on Librispeech}
\label{subsec:reslibrisp}

On the Librispeech dataset, we use the word pieces as the output labels. Since the word pieces are obtained without referring to any acoustic knowledge, the acoustic boundary between adjacent labels may be blurred. Even so, our big CIF-based model still achieves a word error rate (WER) of 2.86\% on test-clean and 8.08\% on test-other (as shown in Table \ref{tab:LibrispeechResults}), which not only shows a clear performance advantage than other soft and monotonic models (e.g. triggered attention \cite{moritz2019triggered}), but also matches or surpasses most of the published results of end-to-end models.

By fine-tuning the trained big model via the chunk-hopping \cite{dong2019self} mechanism, we enables our big model that uses a SAN encoder to support online recognition. As shown in Table \ref{tab:LibrispeechResults}, the online model obtains a WER of 3.25\% on test-clean and 9.63\% on test-other. Besides, the above CIF-based models all apply a very low encoded frame rate (12.5 Hz) for reducing the computational burden. Switching to a higher frame rate may further improve their performance.


\vspace{-13.5pt}
\begin{table}[!ht]
\centering
\caption{Comparison with other end-to-end models on Librispeech, word error rate (WER) (\%)}
\vspace{4.5pt}
\setlength{\tabcolsep}{0.5mm}{
\begin{tabular}{lcccc}
\toprule
& \multicolumn{2}{c}{\textbf{test-clean}} & \multicolumn{2}{c}{\textbf{test-other}} \\
\cmidrule(r){2-3}\cmidrule(r){4-5}
\textbf{Model} & \textbf{w/o LM} & \textbf{w/ LM} & \textbf{w/o LM} & \textbf{w/ LM} \\
\midrule
LAS + SpecAugment \cite{park2019specaugment} & 2.8 & 2.5 & 6.8 & 5.8 \\
Attention + Tsf LM \cite{luscher2019rwth} & 4.4 & 2.8 & 13.5 & 9.3 \\
Jasper \cite{li2019jasper} & 3.86 & 2.95 & 11.95 & 8.79  \\
wav2letter++ \cite{pratap2018wav2letter++} & - & 3.44 & - & 11.24 \\
Cnv Cxt Tsf \cite{mohamed2019transformers} & 4.7 & - & 12.9 & - \\
CTC + SAN \cite{salazar2019self} & - & 4.8 & - & 13.1 \\
CTC + Policy \cite{zhou2018improving} & - & 5.42 & - & 14.70 \\
Triggered Attention \cite{moritz2019triggered} & 7.4 & 5.7 & 19.2 & 16.1 \\
\midrule
CIF + SAN (base) & 4.48 & 3.68 & 12.62 & 10.89 \\
CIF + SAN (big)  & 3.41 & 2.86 & 9.28 & 8.08 \\
\;+ Chunk-hopping (online) & 3.96 & 3.25 & 11.19 & 9.63 \\
\bottomrule
\end{tabular}
}
\vspace{-13.5pt}
\label{tab:LibrispeechResults}
\end{table}

\subsection{Ablation Study on Librispeech}
\label{subsec:ablat}
In this section, we use ablation study to evaluate the importance of different methods applied to the CIF-based model.

\vspace{-13.5pt}
\begin{table}[!ht]
\centering
\caption{Ablation study on Librispeech, where the full model is our base CIF-based model (without LM) in Table.1, WER (\%)}
\vspace{4.5pt}
\begin{tabular}{ccccc}
\toprule
& \textbf{test-clean} & \textbf{test-other} \\
\midrule
without \textbf{scaling strategy} & 6.03 & 14.98 \\
without \textbf{quantity loss} & 8.84 & 15.49 \\
without \textbf{tail handling} & 6.04 & 14.11 \\
without \textbf{CTC loss} & 4.96 & 13.27 \\
without \textbf{autoregressive} & 9.27 & 21.56 \\
\midrule
\textbf{Full Model} & \textbf{4.48} & \textbf{12.62} \\
\bottomrule
\end{tabular}
\label{tab:AblationStudy}
\end{table}

As shown in Table \ref{tab:AblationStudy}, ablating the auto-regression in the decoder causes the largest performance degradation. To further verify this phenomenon, we compare the models with/without auto-regression on the Mandarin dataset of AISHELL-2 but find they achieve comparable performance. Since the acoustic boundaries between Mandarin characters are much more clear, we suspect the importance of auto-regression is related to the clearness of acoustic boundary between output labels. Besides, the proposed support strategies (scaling strategy, quantity loss and tail handling) for the CIF-based model all provide clear improvements. Among them, the quantity loss is the most important since ablating it causes the largest performance loss and brings large learning instability. The introduced CTC loss also benefits to the CIF-based model but not as important as others.

\subsection{Results on AISHELL-2}
\label{subsec:resaishell-2}
On the AISHELL-2 dataset, we use the characters of Mandarin as the output labels. Since every character of Mandarin is single syllable and AISHELL-2 is a read-speech dataset, the acoustic boundary between labels are clear. Consistent with our expectations, the CIF-based model performs very competitive on all test sets and significantly improves the results achieved by the Chain model \cite{povey2016purely}.

\vspace{-13.5pt}
\begin{table}[!ht]
\centering
\caption{Comparison with the previously published results on AISHELL-2, character error rate (CER) (\%)}
\vspace{4.5pt}
\setlength{\tabcolsep}{1.0mm}{
\begin{tabular}{lccc}
\toprule
\textbf{Model} & \textbf{test\_android} & \textbf{test\_ios} & \textbf{test\_mic} \\
\midrule
Chain-TDNN \cite{povey2016purely} & 9.59 & 8.81 & 10.87  \\
\midrule
CIF + SAN & \textbf{6.17} & \textbf{5.78} & \textbf{6.34} \\
\;+ Chunk-hopping (online) & 6.52 & 6.04 & 6.68 \\
\bottomrule
\end{tabular}
}
\vspace{-13.5pt}
\label{tab:AISHELL2Results}
\end{table}

\subsection{Results on HKUST}
\label{subsec:reshkust}

On the HKUST dataset, the speech are all Mandarin telephone conversations, which are more challenging to recognize than read-speech due to the spontaneous and informal speaking style. Besides, the amount of training data on HKUST is smaller. Nevertheless, the CIF-based model still shows good generalization and creates new state-of-the-art result on this benchmark dataset.

\vspace{-13.5pt}
\begin{table}[!ht]
\centering
\caption{Comparison with the previously published results on HKUST, CER (\%)}
\vspace{4.5pt}
\begin{tabular}{lcc}
\toprule
\textbf{Model} & \textbf{CER} \\
\midrule
Chain-TDNN \cite{povey2016purely} & 23.7 \\
Self-attention Aligner \cite{dong2019self} & 24.1 \\
Transformer \cite{zhou2018comparison} & 26.6 \\
Extended-RNA \cite{dong2018extending} & 26.8 \\
Joint CTC-attention model / ESPNet \cite{kim2017joint} & 27.4  \\
Triggered Attention \cite{moritz2019triggered} & 30.5 \\
\midrule
CIF + SAN  & \textbf{23.09} \\
\;+ Chunk-hopping (online) & 23.60 \\
\bottomrule
\end{tabular}
\label{tab:HkustResults}
\end{table}
\vspace{-13.5pt}

\section{DISCUSSION AND CONCLUSION}
\label{sec:concl}
At the theoretical level, CIF simulates the dynamic characteristics of the integrate-and-fire (IF) model on artificial neural networks. The IF model has the dynamics of $I=C*(dU_m/dt)$, where the membrane potential $U_m$ is constantly simulated by the input spikes $I$ in the period of $dt$, $C$ is a constant. Similarly, the dynamics of CIF can be described as $f(h)=d\alpha^a/dt$, which follows the basic dynamic form of the IF model but differs at one aspect: CIF regards the information in the period of $dt$ as a whole and uses continuous values to represent and process. Specifically, CIF uses a vector $h$ to directly represent the inputs in $dt$ and a continuous function f() to directly calculate the change of $\alpha^a$ brought by the inputs. This coarse-grained dynamics determines CIF needs to divide the information when it produces a firing in the period of a encoder step ($dt$). In the future, mimicking the dynamics of other models in spiking neural networks may be a way to improve CIF.

At the application level, CIF not only shows competitive performance on popular ASR benchmarks, but also could extract acoustic embeddings (which may be useful in multimodal tasks, etc.) in a concise way. In addition, CIF could support various sequence transduction tasks (e.g. TTS) by using a suitable encoding resolution. In the future, we will further verify the performance of CIF-based model on larger-scale ASR datasets and other tasks.

\vfill\pagebreak
%
%
\bibliographystyle{IEEEbib}
\bibliography{strings,refs}

\begin{thebibliography}{10}

\bibitem{chorowski2015attention}
Jan~K Chorowski, Dzmitry Bahdanau, Dmitriy Serdyuk, Kyunghyun Cho, and Yoshua
  Bengio,
\newblock ``Attention-based models for speech recognition,''
\newblock in {\em Advances in Neural Information Processing Systems}, 2015.

\bibitem{chan2016listen}
William Chan, Navdeep Jaitly, Quoc Le, and Oriol Vinyals,
\newblock ``Listen, attend and spell: A neural network for large vocabulary
  conversational speech recognition,''
\newblock in {\em ICASSP}, 2016.

\bibitem{prabhavalkar2017comparison}
Rohit Prabhavalkar, Kanishka Rao, Tara~N Sainath, Bo~Li, Leif Johnson, and
  Navdeep Jaitly,
\newblock ``A comparison of sequence-to-sequence models for speech
  recognition.,''
\newblock in {\em INTERSPEECH}, 2017.

\bibitem{chiu2018state}
Chung-Cheng Chiu, Tara~N Sainath, Yonghui Wu, Rohit Prabhavalkar, Patrick
  Nguyen, Zhifeng Chen, Anjuli Kannan, Ron~J Weiss, Kanishka Rao, Ekaterina
  Gonina, et~al.,
\newblock ``State-of-the-art speech recognition with sequence-to-sequence
  models,''
\newblock in {\em ICASSP}, 2018.

\bibitem{lapicque1907recherches}
Louis Lapicque,
\newblock ``Recherches quantitatives sur l'excitation electrique des nerfs
  traitee comme une polarization,''
\newblock {\em Journal de Physiologie et de Pathologie Generalej}, 1907.

\bibitem{abbott1999lapicque}
Larry~F Abbott,
\newblock ``Lapicque's introduction of the integrate-and-fire model neuron
  (1907),''
\newblock {\em Brain research bulletin}, 1999.

\bibitem{Maass1997Networks}
Wolfgang Maass,
\newblock ``Networks of spiking neurons: The third generation of neural network
  models,''
\newblock {\em Neural Networks}, 1997.

\bibitem{hou2017gaussian}
Junfeng Hou, Shiliang Zhang, and Li-Rong Dai,
\newblock ``Gaussian prediction based attention for online end-to-end speech
  recognition.,''
\newblock in {\em INTERSPEECH}, 2017.

\bibitem{tjandra2017local}
Andros Tjandra, Sakriani Sakti, and Satoshi Nakamura,
\newblock ``Local monotonic attention mechanism for end-to-end speech and
  language processing,''
\newblock in {\em Proceedings of the IJCNLP}, 2017.

\bibitem{Merboldt2019}
André Merboldt, Albert Zeyer, Ralf Schlüter, and Hermann Ney,
\newblock ``An analysis of local monotonic attention variants,''
\newblock in {\em INTERSPEECH}, 2019.

\bibitem{chiu2017monotonic}
Chung-Cheng Chiu and Colin Raffel,
\newblock ``Monotonic chunkwise attention,''
\newblock {\em arXiv}, 2017.

\bibitem{fan2018online}
Ruchao Fan, Pan Zhou, Wei Chen, Jia Jia, and Gang Liu,
\newblock ``An online attention-based model for speech recognition,''
\newblock {\em INTERSPEECH}, 2019.

\bibitem{moritz2019triggered}
Niko Moritz, Takaaki Hori, and Jonathan Le~Roux,
\newblock ``Triggered attention for end-to-end speech recognition,''
\newblock in {\em ICASSP}, 2019.

\bibitem{li2019end}
Mohan Li, Min Liu, and Hattori Masanori,
\newblock ``End-to-end speech recognition with adaptive computation steps,''
\newblock in {\em ICASSP}, 2019.

\bibitem{dong2019self}
Linhao Dong, Feng Wang, and Bo~Xu,
\newblock ``Self-attention aligner: A latency-control end-to-end model for asr
  using self-attention network and chunk-hopping,''
\newblock in {\em ICASSP}, 2019.

\bibitem{kim2017joint}
Suyoun Kim, Takaaki Hori, and Shinji Watanabe,
\newblock ``Joint ctc-attention based end-to-end speech recognition using
  multi-task learning,''
\newblock in {\em ICASSP}, 2017.

\bibitem{panayotov2015librispeech}
Vassil Panayotov, Guoguo Chen, Daniel Povey, and Sanjeev Khudanpur,
\newblock ``Librispeech: an asr corpus based on public domain audio books,''
\newblock in {\em ICASSP}, 2015.

\bibitem{du2018aishell}
Jiayu Du, Xingyu Na, Xuechen Liu, and Hui Bu,
\newblock ``Aishell-2: Transforming mandarin asr research into industrial
  scale,''
\newblock {\em arXiv}, 2018.

\bibitem{liu2006hkust}
Yi~Liu, Pascale Fung, Yongsheng Yang, Christopher Cieri, Shudong Huang, and
  David Graff,
\newblock ``Hkust/mts: A very large scale mandarin telephone speech corpus,''
\newblock in {\em Chinese Spoken Language Processing}. 2006.

\bibitem{ko2015audio}
Tom Ko, Vijayaditya Peddinti, Daniel Povey, and Sanjeev Khudanpur,
\newblock ``Audio augmentation for speech recognition,''
\newblock in {\em Sixteenth Annual Conference of the ISCA}, 2015.

\bibitem{park2019specaugment}
Daniel~S Park, William Chan, Yu~Zhang, Chung-Cheng Chiu, Barret Zoph, Ekin~D
  Cubuk, and Quoc~V Le,
\newblock ``Specaugment: A simple data augmentation method for automatic speech
  recognition,''
\newblock {\em INTERSPEECH}, 2019.

\bibitem{sennrich2016neural}
Rico Sennrich, Barry Haddow, and Alexandra Birch,
\newblock ``Neural machine translation of rare words with subword units,''
\newblock in {\em Proceedings of the ACL}, 2016.

\bibitem{abadi2016tensorflow}
Mart{\'\i}n Abadi, Ashish Agarwal, Paul Barham, Eugene Brevdo, Zhifeng Chen,
  Craig Citro, Greg~S Corrado, Andy Davis, Jeffrey Dean, Matthieu Devin,
  et~al.,
\newblock ``Tensorflow: Large-scale machine learning on heterogeneous
  distributed systems,''
\newblock {\em arXiv}, 2016.

\bibitem{ba2016layer}
Jimmy~Lei Ba, Jamie~Ryan Kiros, and Geoffrey~E Hinton,
\newblock ``Layer normalization,''
\newblock {\em arXiv}, 2016.

\bibitem{chorowski2016towards}
Jan Chorowski and Navdeep Jaitly,
\newblock ``Towards better decoding and language model integration in sequence
  to sequence models,''
\newblock {\em INTERSPEECH}, 2017.

\bibitem{bengio2015scheduled}
Samy Bengio, Oriol Vinyals, Navdeep Jaitly, and Noam Shazeer,
\newblock ``Scheduled sampling for sequence prediction with recurrent neural
  networks,''
\newblock in {\em Advances in Neural Information Processing Systems}, 2015.

\bibitem{luscher2019rwth}
Christoph L{\"u}scher, Eugen Beck, Kazuki Irie, Markus Kitza, Wilfried Michel,
  Albert Zeyer, Ralf Schl{\"u}ter, and Hermann Ney,
\newblock ``Rwth asr systems for librispeech: Hybrid vs attention,''
\newblock {\em INTERSPEECH}, 2019.

\bibitem{li2019jasper}
Jason Li, Vitaly Lavrukhin, Boris Ginsburg, Ryan Leary, Oleksii Kuchaiev,
  Jonathan~M Cohen, Huyen Nguyen, and Ravi~Teja Gadde,
\newblock ``Jasper: An end-to-end convolutional neural acoustic model,''
\newblock {\em INTERSPEECH}, 2019.

\bibitem{pratap2018wav2letter++}
Vineel Pratap, Awni Hannun, Qiantong Xu, Jeff Cai, Jacob Kahn, Gabriel
  Synnaeve, Vitaliy Liptchinsky, and Ronan Collobert,
\newblock ``wav2letter++: The fastest open-source speech recognition system,''
\newblock {\em arXiv}, 2018.

\bibitem{mohamed2019transformers}
Abdelrahman Mohamed, Dmytro Okhonko, and Luke Zettlemoyer,
\newblock ``Transformers with convolutional context for asr,''
\newblock {\em arXiv}, 2019.

\bibitem{salazar2019self}
Julian Salazar, Katrin Kirchhoff, and Zhiheng Huang,
\newblock ``Self-attention networks for connectionist temporal classification
  in speech recognition,''
\newblock in {\em ICASSP}, 2019.

\bibitem{zhou2018improving}
Yingbo Zhou, Caiming Xiong, and Richard Socher,
\newblock ``Improving end-to-end speech recognition with policy learning,''
\newblock in {\em ICASSP}, 2018.

\bibitem{povey2016purely}
Daniel Povey, Vijayaditya Peddinti, Daniel Galvez, Pegah Ghahremani, Vimal
  Manohar, Xingyu Na, Yiming Wang, and Sanjeev Khudanpur,
\newblock ``Purely sequence-trained neural networks for asr based on
  lattice-free mmi.,''
\newblock in {\em INTERSPEECH}, 2016.

\bibitem{zhou2018comparison}
Shiyu Zhou, Linhao Dong, Shuang Xu, and Bo~Xu,
\newblock ``A comparison of modeling units in sequence-to-sequence speech
  recognition with the transformer on mandarin chinese,''
\newblock in {\em International Conference on Neural Information Processing},
  2018.

\bibitem{dong2018extending}
Linhao Dong, Shiyu Zhou, Wei Chen, and Bo~Xu,
\newblock ``Extending recurrent neural aligner for streaming end-to-end speech
  recognition in mandarin,''
\newblock {\em INTERSPEECH}, 2018.

\end{thebibliography}

\end{document}